\documentclass[10pt,journal,compsoc]{IEEEtran}
\usepackage{graphicx}
\usepackage{subfigure}

\usepackage{amsmath}
\usepackage{float}
\usepackage{color}

\usepackage{xspace}
\usepackage{multirow}
\usepackage{booktabs}
\usepackage[misc]{ifsym}
\usepackage{amssymb}
\usepackage{ragged2e}

\newcommand{\RNum}[1]{\uppercase\expandafter{\romannumeral #1\relax}}

\usepackage{subfigure}
\usepackage{booktabs}

\usepackage{paralist, tabularx}
\usepackage{multirow} 

\usepackage[colorlinks,linkcolor=blue]{hyperref}

\ifCLASSOPTIONcompsoc
  \usepackage[nocompress]{cite}
\else
  \usepackage{cite}
\fi
\ifCLASSINFOpdf
\else
\fi

\begin{document}

\title{EDFace-Celeb-1M: Benchmarking Face Hallucination with a Million-scale Dataset}

\author{
Kaihao Zhang, Dongxu Li, \ Wenhan Luo,  \ Jingyu Liu,
\ Jiankang Deng, \ Wei Liu, \ Stefanos Zafeiriou \\
\IEEEcompsocitemizethanks{\IEEEcompsocthanksitem K. Zhang and D. Li are with the College of Engineering and Computer Science, Australian National University, Canberra, Australia.\protect\\
E-mail: super.khzhang@gmail.com, dongxu.li@anu.edu.au
\IEEEcompsocthanksitem W. Luo is with Sun Yat-sen University, Guangzhou, China.
E-mail: whluo.china@gmail.com.\protect
\IEEEcompsocthanksitem J. Liu and W. Liu are with the Tencent, Shenzhen, China.
E-mail: jyliu0329@gmail.com; wl2223@columbia.edu.\protect 
\IEEEcompsocthanksitem J. Deng and S. Zafeiriou are with Department of Computing, Imperial College London, London, UK. E-mail: \{j.deng16, s.zafeiriou\}@imperial.ac.uk. \protect
}}

\IEEEtitleabstractindextext{
\begin{abstract}
\justifying  
Recent deep face hallucination methods show stunning performance in super-resolving severely degraded facial images, even surpassing human ability. However, these algorithms are mainly evaluated on non-public synthetic datasets. It is thus unclear how these algorithms perform on public face hallucination datasets. Meanwhile, most of the existing datasets do not well consider the distribution of races, which makes face hallucination methods trained on these datasets biased toward some specific races. 
To address the above two problems, in this paper, we build a public Ethnically Diverse Face dataset, EDFace-Celeb-1M, and design a benchmark task for face hallucination. 
Our dataset includes $1.7$ million photos that cover different countries, with relatively balanced race composition. 
To the best of our knowledge, it is the largest-scale and publicly available face hallucination dataset in the wild. 
Associated with this dataset, this paper also contributes various evaluation protocols and provides comprehensive analysis to benchmark the existing state-of-the-art methods. 
The benchmark evaluations demonstrate the performance and limitations of state-of-the-art algorithms. https://github.com/HDCVLab/EDFace-Celeb-1M

\end{abstract}

}

\maketitle

\IEEEdisplaynontitleabstractindextext

%
\IEEEpeerreviewmaketitle

\section{Introduction}

Human faces contain important identity information and are central to various vision applications, such as face alignment \cite{sun2013deep,zhu2016face,bulat2017far}, face parsing \cite{liu2015multi,lin2019face} and face identification \cite{wang2018cosface,deng2019arcface}. However, most of these applications require high-quality images as input and the approaches perform less favorably in low-resolution conditions. To alleviate the issue, the task of face hallucination, or face super-resolution, aims to super-resolve low-resolution face images to their high-resolution counterparts, thus facilitating effective face analysis.

As a special case of Single Image Super-Resolution (SISR) \cite{dong2014learning,dong2016accelerating,kim2016accurate,lim2017enhanced,lai2017deep,zhang2018residual,zhang2018image,haris2018deep,anwar2020densely,niu2020single}, face hallucination is a fundamental and challenging problem in face analysis. Different from general SISR, which deals with super-resolving pixels in arbitrary scenes, the face hallucination task tackles only facial images. Therefore, the facial special prior knowledge in face images could help recover accurate face shape and rich facial details. As a result, the face hallucination methods often achieve better performance than general single image super-resolution ones in terms of higher up-scaling factors. Previous face hallucination methods \cite{chakrabarti2007super,liu2007face,jia2008generalized,huang2010super,ma2010hallucinating,jung2011position,yang2013structured,wang2014comprehensive,jiang2014face} utilize facial priors to restore high-resolution images by approaches which are typically not end-to-end ones. Currently, a number of deep learning based methods \cite{zhou2015learning,zhu2016deep,huang2017wavelet,cao2017attention,chen2018fsrnet,yu2018face,ma2020deep,jiang2020dual} are proposed to greatly boost the performance of the task of face hallucination, even surpassing human ability.

\textbf{Is face hallucination solved?} Firstly, many current deep hallucination methods evaluate their methods on \textit{non-public} synthetic datasets. Especially, they typically download public datasets and synthesize pairs of low-resolution and high-resolution images, and then \textit{randomly} select some faces as training and testing samples. After that, they evaluate their methods and re-train previous methods on their synthesized datasets. However, using this way for evaluation presents two obvious problems. (1) Given that the division of training/testing groups is random, the following researchers cannot strictly follow and reproduce the previous experiments like \cite{yu2016ultra,yu2018super}. (2) To verify that the new face hallucination methods outperform the previous methods, researchers have to synthesize new datasets by themselves and re-train previous methods again, which greatly increases the unnecessary workload and reduces the credibility of their results like \cite{huang2017wavelet,ma2020deep}. Therefore, it is not clear how these algorithms would perform on public face hallucination datasets.

On the other hand, though it is popular that the current deep face hallucination methods are evaluated on synthesized face datasets, these datasets face the problem of being biased toward specific races, and other races are significantly ignored. This scheme along with these datasets not only fails to accurately evaluate the performance of face hallucination methods, but also will raise the ethical problem. Therefore, a large-scale face hallucination dataset with relatively balanced race composition is necessary for face analysis in the community.

To address the above problems, in this paper, we introduce an Ethnically Diverse Face dataset called EDFace-Celeb-1M, and design benchmark protocols along with analysis to evaluate and encourage the development of face hallucination algorithms. There exist three key objectives of creating the EDFace-Celeb-1M dataset for face hallucination. (1) It should contain a large-scale set of face images in the wild with unconstrained pose, emotion and exposure. Specially, the proposed EDFace-Celeb-1M dataset includes $1.7$ million photos of more than $40,000$ unique celebrity subjects. (2) The dataset should include faces from as many different countries and races as possible to mitigate the race bias in the current face hallucination datasets. More specifically, the EDFace-Celeb-1M dataset contains different race groups including White, Black, Latino and Asian with a relatively balanced composition. (3) It should be publicly available, enabling benchmarking current and future face hallucination methods with a unified dataset protocol to ensure fair and effortless evaluation.  In this paper, the proposed dataset is public available with fixed training and testing sets for fair comparison.

To investigate the performance of current deep face hallucination algorithms on the constructed dataset with relatively balanced race composition, we design and provide concrete evaluation protocols, and evaluate four publicly available face hallucination methods and four SISR methods. We also introduce in detail our experiment setup and report baseline models to benefit and drive future research for the face hallucination task and inspire other related tasks in the computer vision community.

Our main contributions are summarized as follows.

\begin{figure*}
  \centering
  \includegraphics[width=0.9\linewidth ]{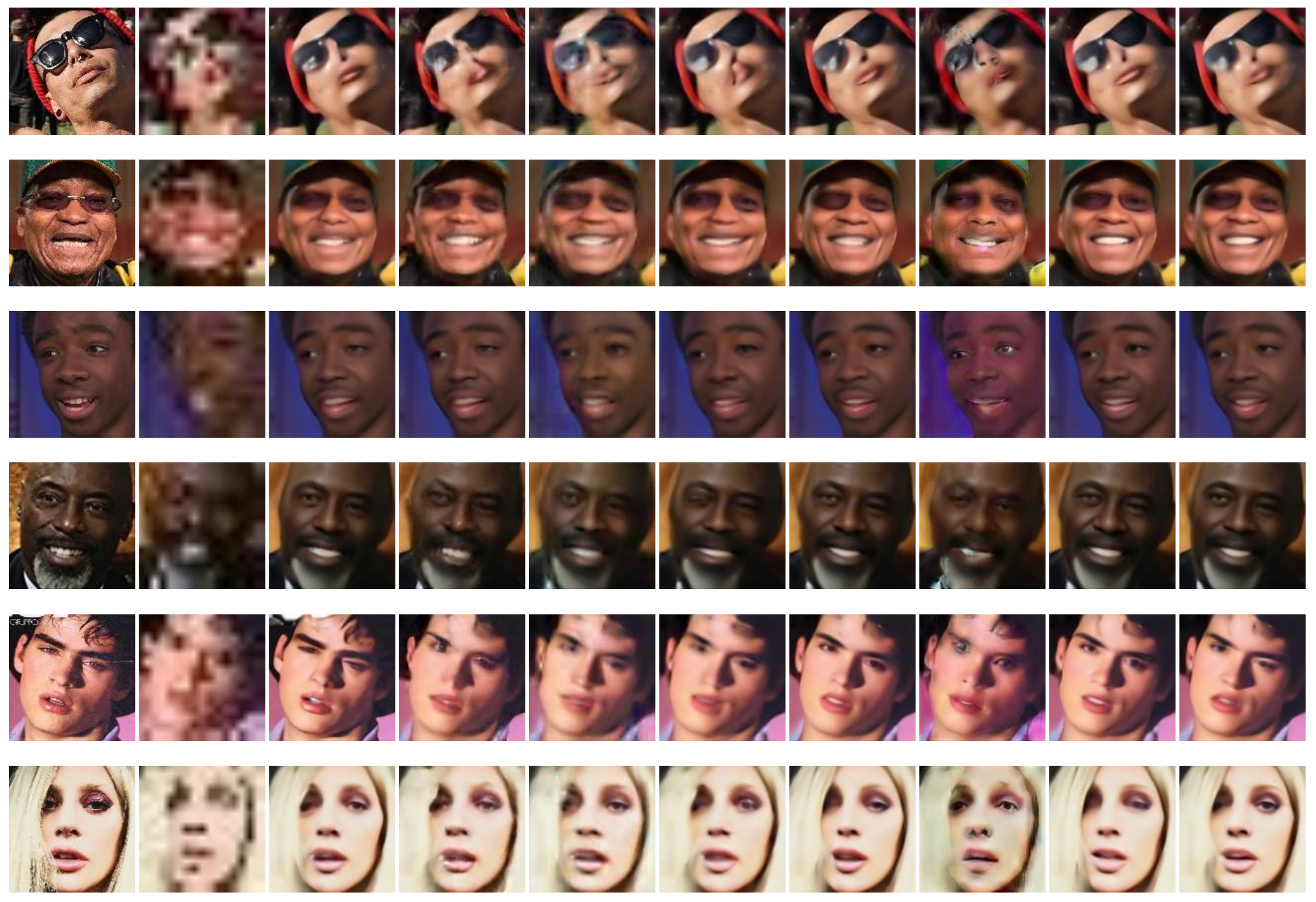}
  \vspace{-4mm}
  \caption{\textbf{Visual results of \textbf{BI} models ($\times 8$) on the EDFace-Celeb-1M dataset}. From left to right: HR, results of bicubic, DICNet, DICGAN, WaveletSR, HiFaceGAN, EDSR, RDN, RCAN and HAN. ``BI" means the bicubic interpolation.}
  \vspace{-4mm}
  \label{fig:results_8x} 
\end{figure*}

\begin{figure*}
  \centering
  \includegraphics[width=0.9\linewidth ]{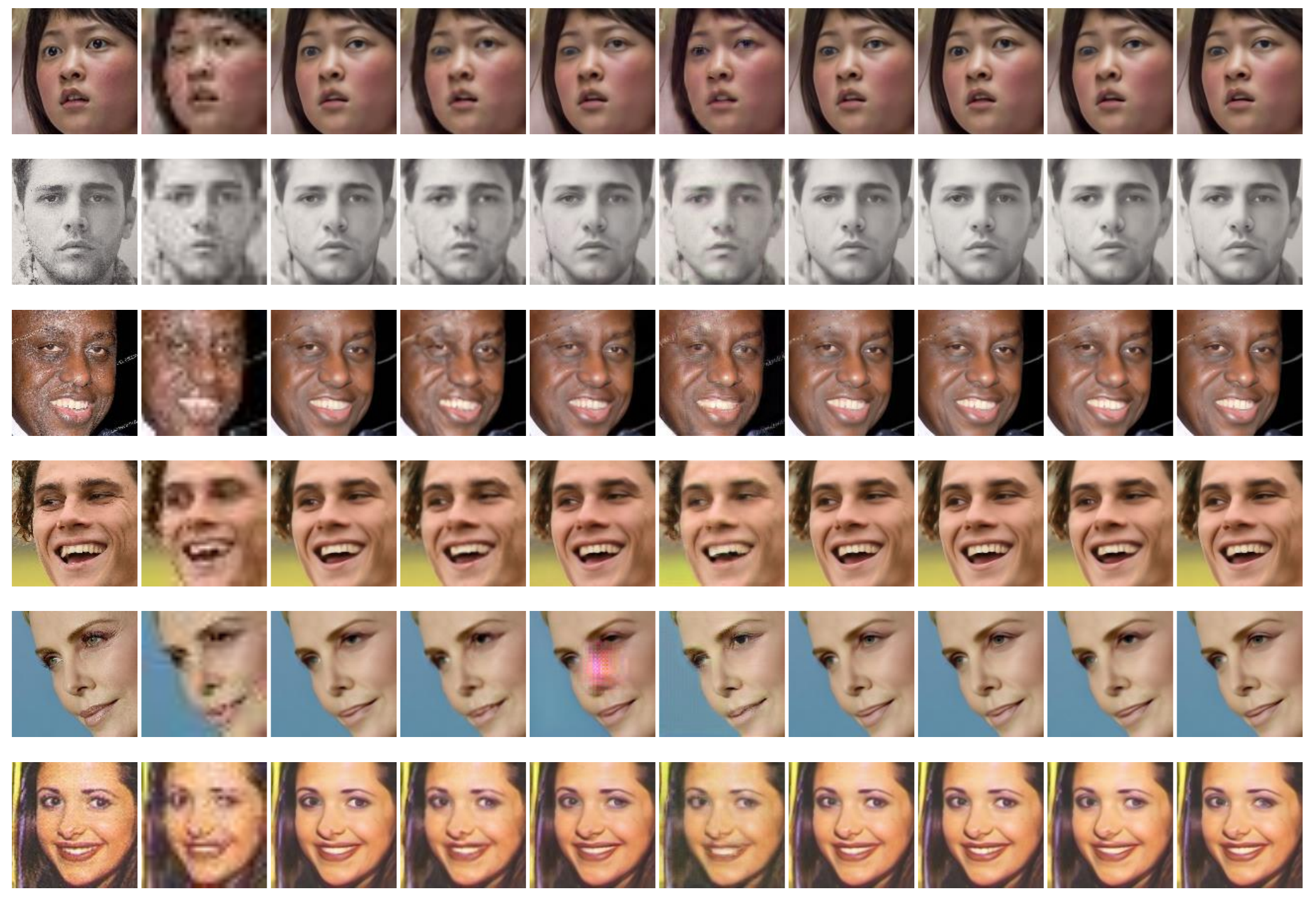}
  \vspace{-4mm}
  \caption{\textbf{Visual results of \textbf{BI} models ($\times 4$) on the EDFace-Celeb-1M dataset}. From left to right: HR, results of bicubic, DICNet, DICGAN, WaveletSR, HiFaceGAN, EDSR, RDN, RCAN and HAN. ``BI" means the bicubic interpolation.}
  \vspace{-4mm}
  \label{fig:results_4x} 
\end{figure*}

\begin{figure*}
  \centering
  \includegraphics[width=0.9\linewidth ]{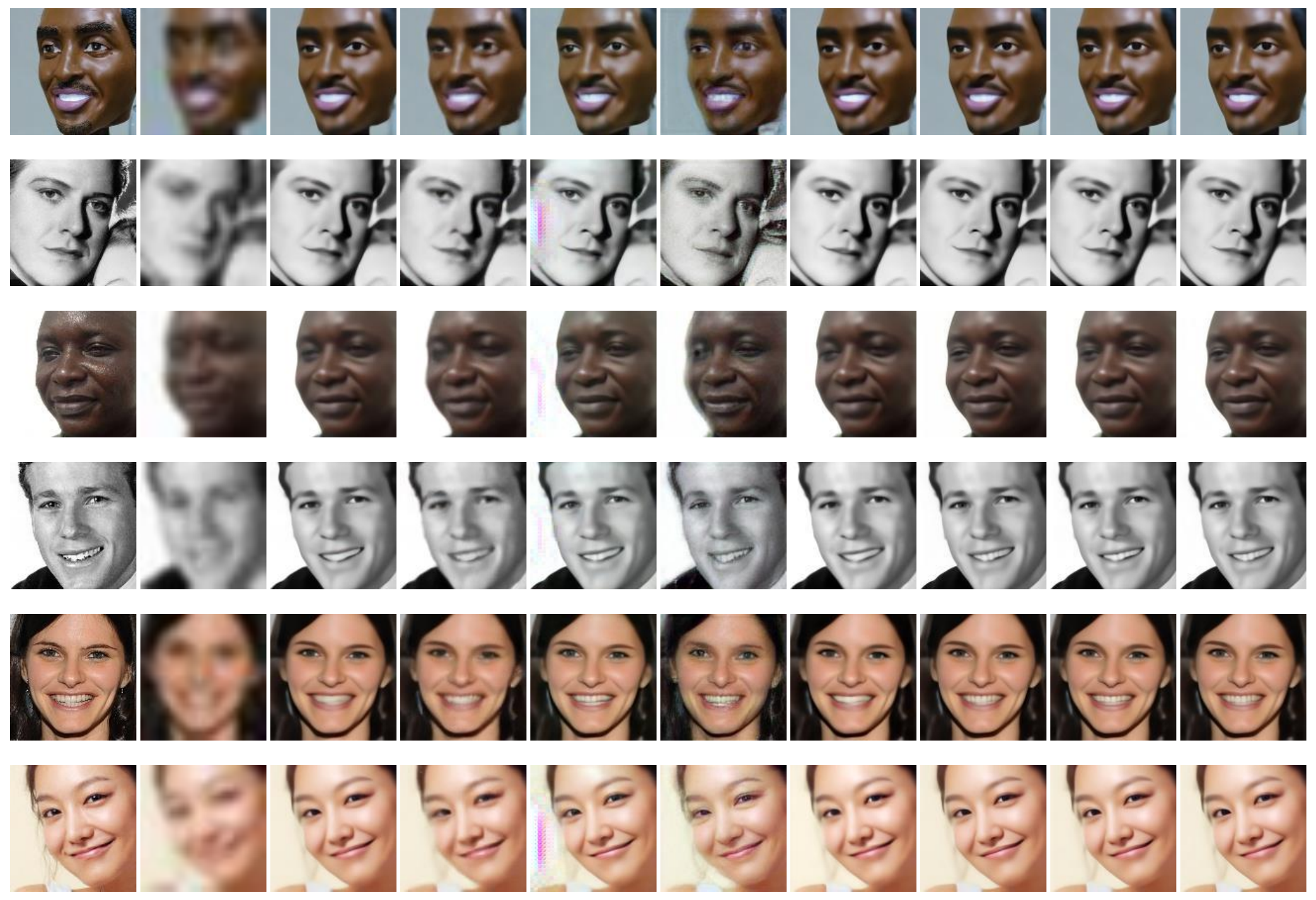}
  \vspace{-4mm}
  \caption{\textbf{Visual results of \textbf{BD} models ($\times 4$) on the EDFace-Celeb-1M dataset}. From left to right: HR, results of bicubic, DICNet, DICGAN, WaveletSR, HiFaceGAN, EDSR, RDN, RCAN and HAN. ``BD" means that the blur artifact is applied prior to the downsampling operation.}
  \vspace{-4mm}
  \label{fig:results_bd} 
\end{figure*}

\begin{figure*}
  \centering
  \includegraphics[width=0.9\linewidth ]{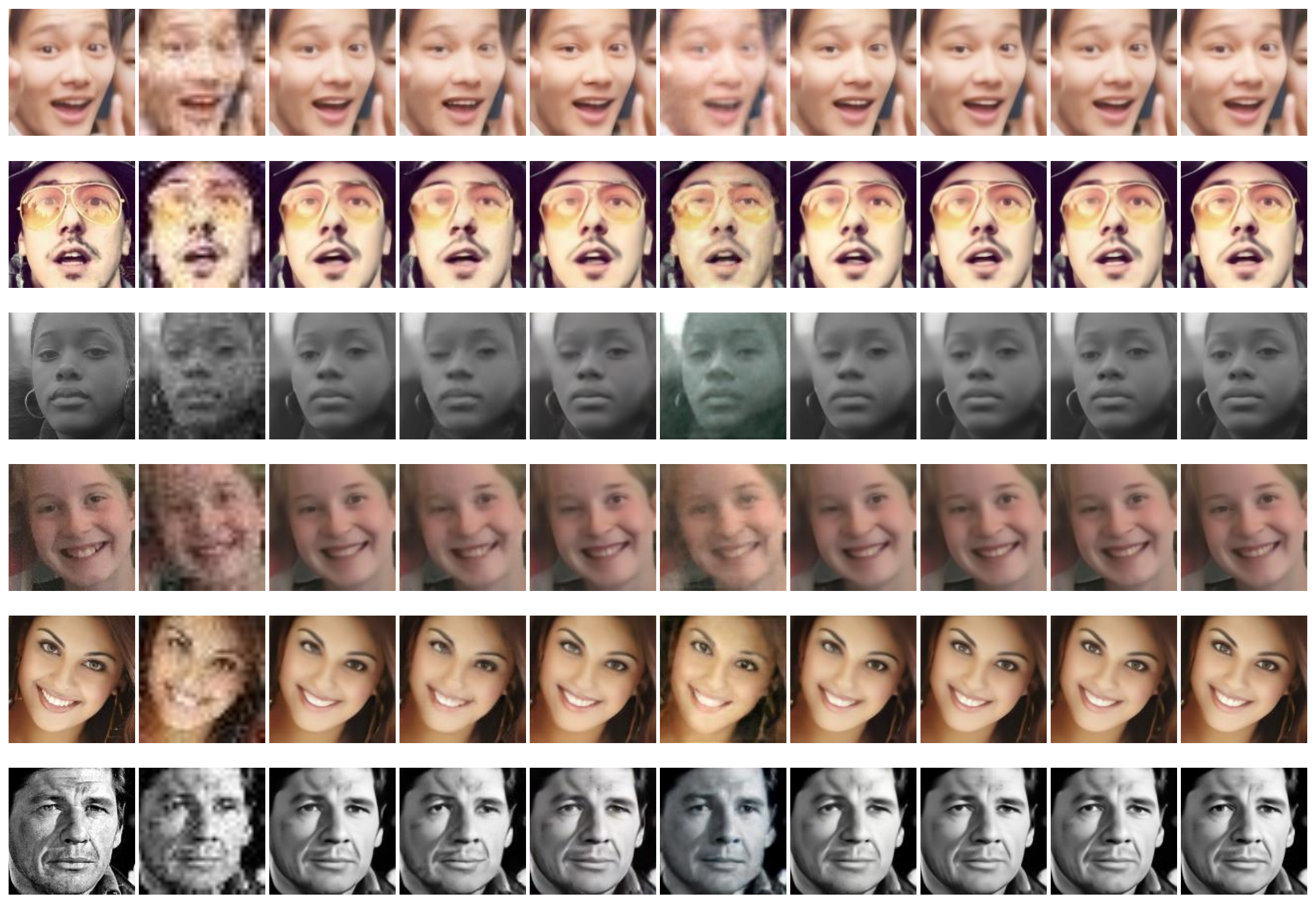}
  \vspace{-4mm}
  \caption{\textbf{Visual results of \textbf{DN} models ($\times 4$) on the EDFace-Celeb-1M dataset}. From left to right: HR, results of bicubic, DICNet, DICGAN, WaveletSR, HiFaceGAN, EDSR, RDN, RCAN and HAN. ``DN" means that Gaussian noise is added to the LR images after the downsampling operation.}
  \vspace{-4mm}
  \label{fig:results_dn} 
\end{figure*}

\begin{itemize}
\item
First, to the best of our knowledge, we build the \textit{first large-scale publicly available} face hallucination dataset with relatively balanced race composition. The dataset includes $1.7$ million face images collected from different race groups, providing \textit{fixed} training and testing groups, pairs of low-resolution and high-resolution images with different scale factors (\textit{e.g., 2$\times$, 4$\times$, 8$\times$}), and aligned and non-aligned face images, which makes the future comparison more convenient, repeatable and credible.
\item
Second, we design a \textit{benchmark} task to evaluate the performance of some current deep face hallucination and SISR methods to super-resolve low-resolution images. By doing so, it is clear how these algorithms perform on public face hallucination datasets (See Section \ref{sec:experiment}).
\item
Third, we address \textit{fundamental questions} of face hallucination and obtain several key findings. 

\begin{itemize}

\item
How well do the current face hallucination and SISR methods perform in the case of different upsampling factors? (See Table \ref{tab:psnr_ssim} and Figure \ref{fig:results_8x},\ref{fig:results_4x},\ref{fig:results_bd},\ref{fig:results_dn})

\item
How do the noise and blur kernels affect the performance of face hallucination methods? (See Table \ref{tab:psnr_ssim} and Figure \ref{fig:results_8x},\ref{fig:results_4x},\ref{fig:results_bd},\ref{fig:results_dn})

\item
Is the size of training data important? (See Table \ref{tab:different_samples} and Figure \ref{fig:different_samples})

\item
Using the super-resolved images for the facial analysis tasks like landmark detection and identity preservation, is the gap significant, compared with using the ground truth high-resolution face images? (See Table \ref{tab:alignment_parsing_id})

\end{itemize}

\end{itemize}

\section{Related Work}

\subsection{Face Hallucination}

Many approaches have been proposed for face hallucination, which can be classified into two categories: non-deep learning methods and deep learning methods. For the non-deep learning methods, holistic-based \cite{wang2005hallucinating,liu2007face,kolouri2015transport} and part-based \cite{ma2010hallucinating,tappen2012bayesian,yang2018hallucinating} techniques are two popular models, which upsample face images via representing faces by parameters and extracting facial regions, respectively.

Recently, deep neural networks have been successfully applied to various computer vision tasks including face hallucination. Yu \textit{et al.} \cite{yu2016ultra} investigate the Generative Adversarial Network (GAN) to super-resolve face images of very low resolution and create perceptually realistic high-resolution face images. Huang \textit{et al.} \cite{huang2017wavelet} introduce wavelet coefficients prediction into deep networks to generate super-resolution face images with different upscaling factors. To train deep face hallucination networks, Zhang \textit{et al.} \cite{zhang2018super} propose a super-identify loss function to measure the difference of identity information. Cao \textit{et al.} \cite{cao2017attention} design a novel attention-aware face hallucination framework and use deep reinforcement learning to optimize its parameters.

As a domain-specific super-resolution problem, there are also many face hallucination methods that use facial prior knowledge to help super-resolve low-resolution face images. Song \textit{et al.} \cite{song2017learning} propose a two-stage framework, which firstly generates facial components to represent the basic facial structures and then synthesizes fine-grained facial structures through a component enhancement method. Yu \textit{et al.} \cite{yu2018face} present a multi-task upsampling network to employ the image appearance similarity and exploit the face structure information with the help of the proposed facial component heat maps. Chen \textit{et al.} \cite{chen2018fsrnet} introduce a FSRNet model to make use of facial landmark heat maps and parsing maps. In addition, the attention mechanism \cite{kim2019progressive} and bi-network \cite{zhu2016deep} are also applied to make use of facial prior knowledge to train a high-resolution face generator.

\subsection{Single Image Super-Resolution}

The advancement of deep neural networks has achieved great success on image super-resolution, and most of the state-of-the-art SISR methods are based on deep learning \cite{wang2020deep}. As a pioneer work of deep SISR methods, Dong \textit{et al.} \cite{dong2014learning, dong2015image} propose a Super-Resolution Convolutional Neural Network (SRCNN) to super-resolve low-resolution images by firstly adopting deep learning for SISR. After that, many improvements have been explored. For example, Kim et al. \cite{kim2016deeply} propose a deeply-recursive CNN to make use of skip connections to train their proposed a Deeply-Recursive Convolutional Network (DRCN). Lim \textit{et al.} \cite{lim2017enhanced} design an Enhanced Deep Super-Resolution Network (EDSR) to remove redundant modules and combine with multi-scale processing. To reduce the computational cost, many efficient SISR methods are proposed \cite{lai2017deep,kim2016accurate,dong2016accelerating}. To make the generated images more realistic, GAN based SISR methods are introduced to improve the perceptual quality of HR images \cite{ledig2017photo,wang2018esrgan,sajjadi2017enhancenet}. Recently, Hairs \textit{et al.} \cite{haris2018deep} develop a Deep Back-Project Network (DBPN) to exploit the mutual dependencies with a feedback mechanism. Zhang \textit{et al.} \cite{zhang2018residual} introduce dense connections to make use of cues. Residual Channel Attention Networks (RCAN) is proposed in \cite{zhang2018image} to introduce a residual-in-residual structure and a channel attention module. More recently, there are many SISR methods utilizing novel attention modules \cite{dai2019second,niu2020single} or feedback mechanisms \cite{li2019feedback} to further improve the performance of SISR. Even without considering facial structures, these above methods can also super-resolve low-resolution face images to their corresponding high-resolution versions. 

\subsection{Face Datasets}

Currently, there does not exist publicly available face hallucination datasets. In this section, we briefly review some popular face datasets which have been constructed recently. The Labeled Faces in the Wild (LFW) dataset \cite{huang2008labeled} is created in 2007, and it contains $13,000$ images. In 2014, the CelebA \cite{sun2014deep} and CASIA-WebFace \cite{yi2014learning} datasets are released, including about $20$K and $500$K images, respectively. The VGGface \cite{parkhi2015deep} dataset released in 2015 includes $2.6$ million images.

More recently, Kemelmacher-Shlizerman \textit{et al.} \cite{kemelmacher2016megaface} assemble a dataset of $4.7$ million images to evaluate how face recognition algorithms perform with a very large number of images. Cao \textit{et al.} \cite{cao2018vggface2} release the VGGface2 dataset. Compared to the VGGface dataset, VGGface2 has $3.3$ million images to cover a larger number of identities. The largest-scale face dataset is MS-Celeb-1M, which contains $10$ million images for training and testing.

Even though there exist many face datasets, none of them can be directly utilized to evaluate the current face hallucination approaches, due to the following reasons. Firstly, none of these datasets provides large-scale pairs of low-resolution and high-resolution face images. However, the current deep face hallucination methods mainly rely on supervised learning and thus pairs of training face images are necessarily required. Secondly, researchers in the computer vision community have paid increasing attention to the race bias problem. However, these existing datasets are strongly biased toward specific races. Face hallucination models trained on these datasets will generate high-resolution face images with inappropriate race information.

The FairFace\cite{karkkainen2019fairface} contains face images from different races. However, it includes only $10$K images without pairs of face images for the evaluation of face hallucination methods. A summary of current face datasets is listed in Table \ref{table_summary} to give a clear view.

\begin{table}[tb]
  \centering 
    \caption{\textbf{Representative face datasets.} Most of the current public face datasets do not consider the race problem. Meanwhile, none of them provides large-scale pairs of LR and HR samples for evaluating deep hallucination methods.} \small
    \begin{tabular}{c | c c c c }
    \toprule
    Dataset & Size &  Public & Race & HR-LR \\
    \midrule
    LFW & 13K & $\checkmark$ & - & $\times$ \\
    CelebA & 200K & $\checkmark$ & - & $\times$ \\  
    CASIA-WebFace & 500K & $\checkmark$ & - & $\times$ \\
    FB-DeepFace & 4.4M & $\times$ & - & $\times$ \\
    VGGFace & 2.6M & $\checkmark$ & - & $\times$ \\
    VGGFace2 & 3.3M & $\checkmark$ & - & $\times$ \\
    UMDFaces & 367K & $\checkmark$ & - & $\times$ \\
    FaceScrub & 100K & $\checkmark$ & - & $\times$ \\
    MegaFace & 1M & $\checkmark$ & - & $\times$ \\ 
    FairFace & 10K & $\checkmark$ & $\checkmark$ & $\times$ \\
    MS-Celeb-1M & 10M & $\checkmark$ & - & $\times$ \\
    \toprule
    \textbf{Ours} & \textbf{1.7M} & \textbf{$\checkmark$} & \textbf{$\checkmark$} & \textbf{$\checkmark$} \\
    \bottomrule
    \end{tabular}%
    \label{table_summary}
\end{table}%

To overcome the problem of lacking face hallucination datasets, current face hallucination researchers synthesize pairs of training and testing samples to evaluate the previous methods and their proposed ones. For example, Yu \textit{et al.} \cite{yu2018super,yu2018face} and Kim \textit{et al.} \cite{kim2019progressive} train and test their proposed methods on pairs of face images synthesized from CelebA \cite{liu2015deep}. Zhang \textit{et al.} \cite{zhang2020copy} synthesize pairs of face images from Multi-PIE \cite{gross2010multi} and CelebA \cite{liu2015deep}. WaveletSR \cite{huang2017wavelet} is evaluated on synthesized face images from VGGface2 \cite{cao2018vggface2}, and Li \textit{et al.} \cite{li2020blind} train and test their methods on the FFHQ \cite{karras2019style}, VGGface2 \cite{cao2018vggface2} and CelebA \cite{liu2015deep} datasets. However, most of these synthesized face hallucination datasets are not public. Meanwhile, given that some of them are randomly split training and testing samples, the face hallucination community faces the following challenge: all the following methods cannot directly use the pre-trained models and their results like (Peak Signal-to-Noise Ratio (PSNR )) and (Structural Similarity Index (SSIM)) values, and thus have to synthesize datasets by themselves again and re-train previous face hallucination methods. Therefore, it increases a lot of unnecessary duplication of works. In order to benchmark current deep face hallucination methods and provide convenience to the future face hallucination research, we provide a large-scale publicly available face hallucination dataset with relatively balanced race composition and contribute several baseline models in this paper.

\begin{figure}[tb]
  \centering
\includegraphics[width=0.9\linewidth ]{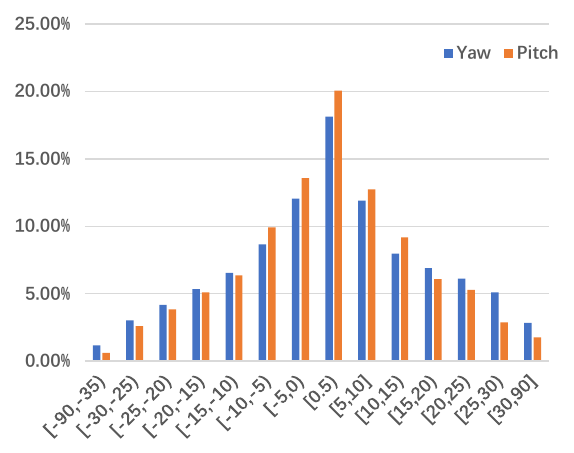}
\caption{{\bf The facial pose statistics of the proposed EDFace-Celeb-1M dataset}.}
\label{face_pose}
\end{figure}

\begin{figure}[tb]
  \centering
\includegraphics[width=0.99\linewidth ]{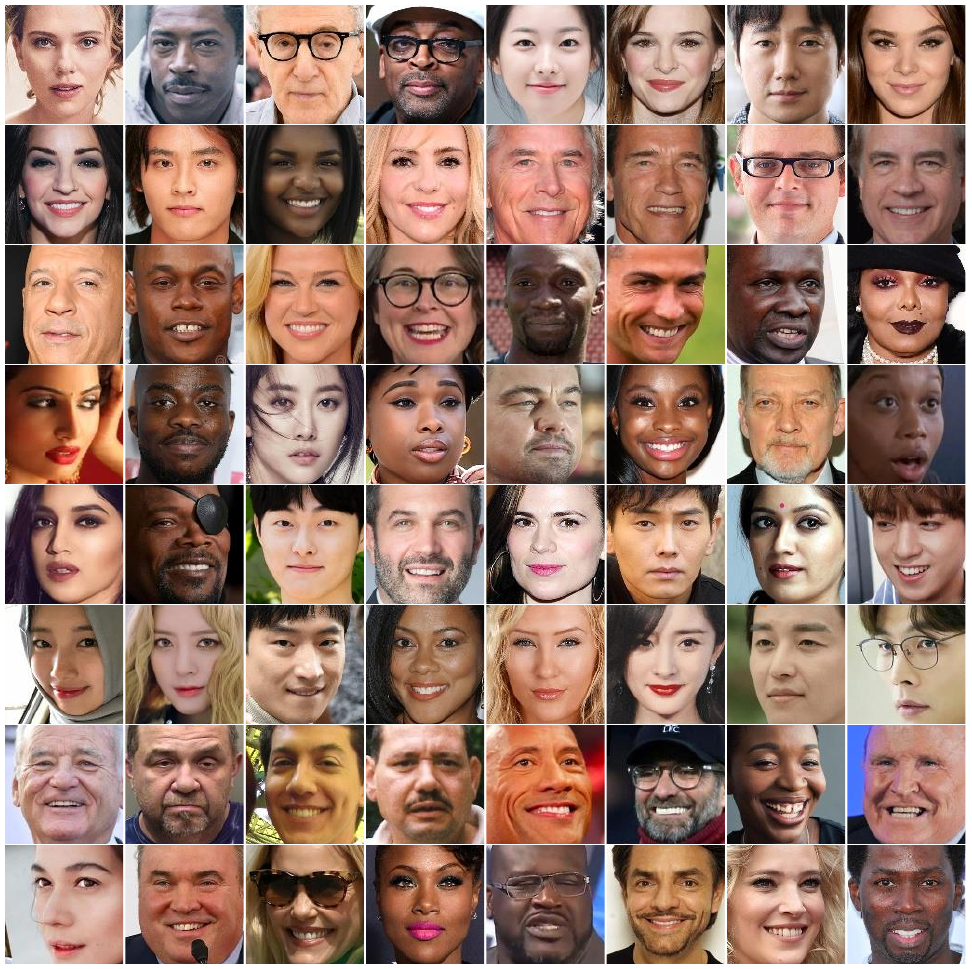}
\caption{{\bf Representative face images from the proposed dataset.} These face images exhibit relatively balanced race distribution}, evident pose and appearance variations.
\label{representative_image}
\end{figure}

\begin{figure}[tb]
  \centering
\includegraphics[width=0.9\linewidth ]{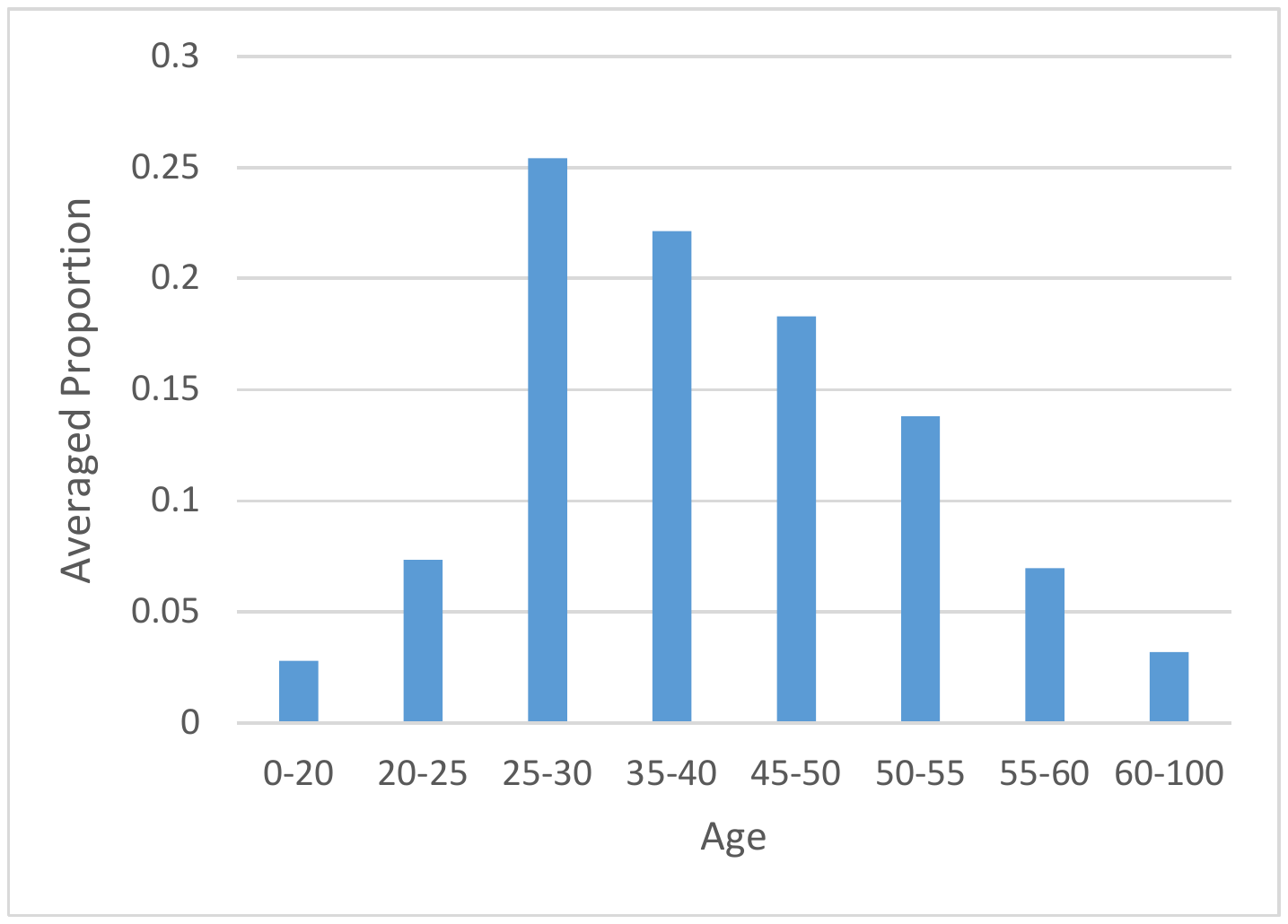}
\caption{{\bf The age statistics of the proposed EDFace-Celeb-1M dataset}.}
\label{age}
\end{figure}

\begin{figure}[tb]
  \centering
\includegraphics[width=0.85\linewidth ]{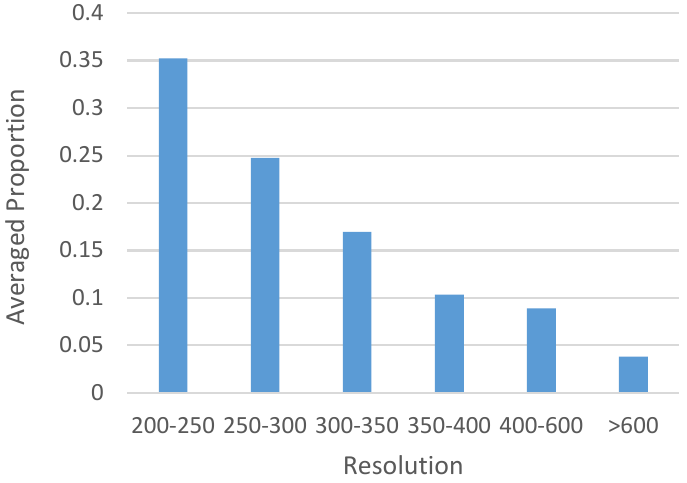}
\caption{{\bf The resolution of face images before resizing}.}
\label{resolution}
\end{figure}

\section{EDFace-Celeb-1M Collection}
\label{sec:collection}

In this section, we provide an overview of the EDFace-Celeb-1M dataset and introduce how it is collected in detail. We build the dataset to benchmark the current deep face hallucination methods and drive the development of the face hallucination task in the future. As mentioned above, we aim to build a publicly available large-scale face hallucination dataset, which provides pairs of low-resolution and high-resolution face images with a fixed setting of training and testing samples. 

The dataset collection processing includes: how a list of candidate identities is obtained, how the candidate images are collected, how to detect the face in images, and how to synthesize pairs of low-resolution and high-resolution images. In addition, we provide attribute statistics of the proposed dataset such as race and gender.

\subsection{Stage \RNum{1}: Obtain a Name List}
The first step from scratch is to have a list of subject names whose faces we aim to collect.
As mentioned before, race balance is the top priority when we build this dataset. To obtain such a name list with race balance, we split the races into four groups: White, Black, Asian and Latino. Based on this, we collect names for different groups. More specifically, for each group, we collect as many names of celebrities as possible from different countries. The names in our list are from diverse countries. For example, the Asian group includes names of people from East Asian, Southeast Asian, Middle East and India from Asian countries. In addition, it also includes some Asian-Americans. In summary, our EDFace-Celeb-1M dataset contains more than $20,000$ names and the ratio (name lists) of the above four groups is about $31.1\% : 19.2\% : 19.6\% : 18.3\%$.

\subsection{Stage \RNum{2}: Select Images for Each Identity}
After obtaining a name list, we use the Google Image Search engine to download $100 \sim 1000$ images for each identity. Moreover, to obtain diverse images with age variations, we further add the keyword young to each subject and further download the corresponding images.

\subsection{Stage \RNum{3}: Face Detection}
We then detect faces in images via the Dlib detector \cite{Dlib}. In this way, we can obtain a facial image dataset, containing faces of different poses/angles, variations of appearance (like glasses, hat). In addition, we manually remove some non-face images.

\subsection{Stage \RNum{4}: Synthesize Pairs of Images}
With the above steps, we have obtained about $10$M facial images in total. Given the obtained facial images, we construct two subsets. The first one is composed of real-world low resolution facial images, which are dedicated to the qualitative study of existing face hallucination methods, as there is no ground truth available. Specifically, we choose images whose resolution is smaller than $50 \times  50$ to compose this subset. 

The second set is dedicated to quantitatively evaluating the existing face hallucination methods, consisting of pairs of high-resolution and low-resolution facial images. To this end, we have to choose high-resolution images and synthesize the corresponding low-resolution images. To be specific, we choose $1.5$M face images whose resolution is larger than $128 \times  128$ and resize them to $128 \times  128$. These images serve as high-resolution images. To synthesize the corresponding low-resolution images, we employ the strategies employed in most of the existing face hallucination methods. More specifically, we simulate the degradation process via specific operations like downsampling. The developed subset includes five different degradation settings, named as \textit{$2\times$}, \textit{$4\times$}, \textit{$4\times$\_BD}, \textit{$4\times$\_DN} and \textit{$8\times$}. The numbers indicate the downsampling factor, ``D" stands for downsampling, ``B" indicates blur operation, and ``N " stands for Gaussian noise that is added to the LR images. The order of letters indicates the order of operations. For example, ``BD" means that the blur artifact is applied prior to the downsampling operation. We use bicubic interpolation for downsampling. $2\times$, $4\times$ and $8\times$ mean that only bicubic downsampling operation is applied. When creating blurry and noisy face images, Gaussian blur and Gaussian noise are added to images. Among the $1.5$M image pairs, we choose $1.36$M pairs of images for training and $0.14$M pairs of images for quantitative testing.

In summary, by conducting the above four steps, we derive a dataset including two subsets of non-aligned facial images. The first one
contains $200$K real-world low-resolution images for qualitative testing, and the second one includes $1.36$M and $140$K pairs of images for training and quantitative testing.

\begin{table*}[tb]

  \centering 
    \caption{\textbf{Performance comparison of representative methods for face hallucination on the EDFace-Celeb-1M dataset}. Results are reported in terms of both PSNR and SSIM.  The highest, second highest and lowest results are highlighted in bolded, blue and red, respectively.}\small
    \resizebox{\textwidth}{!}{
    \begin{tabular}{c | c | c c c c c c c c c }
    \hline
    Scale &  Metrics  & DICNet \cite{ma2020deep} & DICGAN \cite{ma2020deep} & WaveletSR \cite{huang2017wavelet} & HiFaceGAN \cite{yang2020hifacegan} & EDSR \cite{lim2017enhanced} & RDN \cite{zhang2018residual} & RCAN \cite{zhang2018image} & HAN \cite{niu2020single} \\
    \hline
    \multirow{2}{*}{$2\times$} & PSNR & - & - & 30.60 & \textcolor{red}{29.68} & 31.23 & 31.39 & \textbf{31.42} & \textcolor{blue}{31.41} \\
                        & SSIM & - & - & \textbf{0.9119}  & \textcolor{red}{0.8836}  & 0.8869 & 0.8889 & \textcolor{blue}{0.8892} & 0.8888\\
    \hline
    \multirow{2}{*}{$4\times$} & PSNR & \textbf{29.06} & \textcolor{blue}{28.41} & 26.35 & \textcolor{red}{25.37}  & 26.99 & 27.56  & 27.64 & 27.62\\
                        & SSIM & \textbf{0.8453} & \textcolor{blue}{0.8261} & 0.8211  & \textcolor{red}{0.7727}  & 0.8035 & 0.8153 &  0.8161 & 0.8168 \\
    \hline
    \multirow{2}{*}{$4\times\_BD$} & PSNR & \textbf{29.68} & \textcolor{blue}{28.58} & 26.52 & \textcolor{red}{24.23}  & 27.22 & 27.83 & 27.89 & 27.88\\
                        & SSIM & \textbf{0.8213} & \textcolor{blue}{0.7988} & 0.7940 & \textcolor{red}{0.7009} & 0.7825 & 0.7955 & 0.7969 & 0.7964 \\
    \hline
    \multirow{2}{*}{$4\times\_DN$} & PSNR & \textbf{27.96}  & \textcolor{blue}{27.38} & 25.72 & \textcolor{red}{22.94}  & 26.08  & 26.59  &  26.66 & 26.62  \\
                        & SSIM & \textbf{0.8117} & \textcolor{blue}{0.7922} & 0.7815 & \textcolor{red}{0.6856} & 0.7673 & 0.7817 & 0.7835 & 0.7851\\
    \hline
    \multirow{2}{*}{$8\times$} & PSNR & \textbf{25.29} & \textcolor{blue}{24.64} & 22.33 & \textcolor{red}{21.88}  & 23.24 & 23.74 & 23.77 & 23.73 \\
                        & SSIM & \textbf{0.7453}  & \textcolor{blue}{0.7134} & 0.6758 & \textcolor{red}{0.6408} & 0.6890 & 0.7117 & 0.7114 & 0.7114\\
    \bottomrule
    \end{tabular}}%
    \label{tab:psnr_ssim}
\end{table*}%

\begin{table*}[tb]
  \centering 
    \caption{\textbf{Performance comparison of representative methods for face hallucination on the EDFace-Celeb-1M dataset}. Results are reported in terms of the errors of face alignment, face parsing and identity information.  The highest, second highest and lowest results are highlighted in bolded, blue and red, respectively.}\small
    \resizebox{\textwidth}{!}{
    \begin{tabular}{c | c | c c c c c c c c c  }
    \hline
    Scale &  Metrics  & DICNet \cite{ma2020deep} & DICGAN \cite{ma2020deep} & WaveletSR \cite{huang2017wavelet} & HiFaceGAN \cite{yang2020hifacegan} & EDSR \cite{lim2017enhanced} & RDN \cite{zhang2018residual} & RCAN \cite{zhang2018image} & HAN \cite{niu2020single} \\
    \hline
    \multirow{2}{*}{$2\times$} & Alignment & - & - & \textcolor{blue}{0.0218} & \textcolor{red}{0.0223} & 0.0220 & 0.0219 & 0.0220  & \textbf{0.0210} \\
                     & Identity & - & - & \textbf{0.9813} & \textcolor{red}{0.9773} & 0.9800 & 0.9811 & 0.9812 & \textbf{0.9813} \\
    \hline
    \multirow{2}{*}{$4\times$} & Alignment & 0.0259  & 0.0272 & 0.0262 & \textcolor{red}{0.0295} & 0.0264 & 0.0254 & \textcolor{blue}{0.0253} & \textbf{0.0250} \\
                     & Identity & 0.8360 & 0.8184 & 0.8338 & \textcolor{red}{0.7875} & 0.8262 & 0.8490 & \textbf{0.8543} & \textcolor{blue}{0.8527} \\
    \hline
    \multirow{2}{*}{$4\times\_BD$} & Alignment & 0.0257 & 0.0277 &  0.0258 & \textcolor{red}{0.0312} & 0.0261  & 0.0252 & \textbf{0.0250} & \textcolor{blue}{0.0251} \\
                     & Identity & 0.7846 & 0.7726 & 0.7055 & \textcolor{red}{0.6758} & 0.7786 & 0.8045 & \textbf{0.8109} &  \textcolor{blue}{0.8087} \\
    \hline
    \multirow{2}{*}{$4\times\_DN$} & Alignment & 0.0289 & 0.0312 & 0.0292  & \textcolor{red}{0.0362} & 0.0298 & 0.0281 & \textcolor{blue}{0.0280} & \textbf{0.0279} \\
                     & Identity & 0.7035 & 0.6797 & 0.7857 & \textcolor{red}{0.5344} & 0.6969 & 0.7291 & \textbf{0.7332} & \textcolor{blue}{0.7302} \\
    \hline
    \multirow{2}{*}{$8\times$} & Alignment & 0.0366 & 0.040 & 0.0407 & \textcolor{red}{0.0440} & 0.0399 & \textbf{0.0353}  &  \textcolor{blue}{0.0354} & \textcolor{blue}{0.0354} \\
                     & Identity & 0.4830 & 0.448 & 0.4305 & \textcolor{red}{0.3851} & 0.4600 & 0.5094 & \textbf{0.5116}  & \textcolor{blue}{0.5081} \\
    \bottomrule
    \end{tabular}}
    \label{tab:alignment_parsing_id}
\end{table*}%

\begin{figure*}
  \centering
  \includegraphics[width=0.9 \linewidth ]{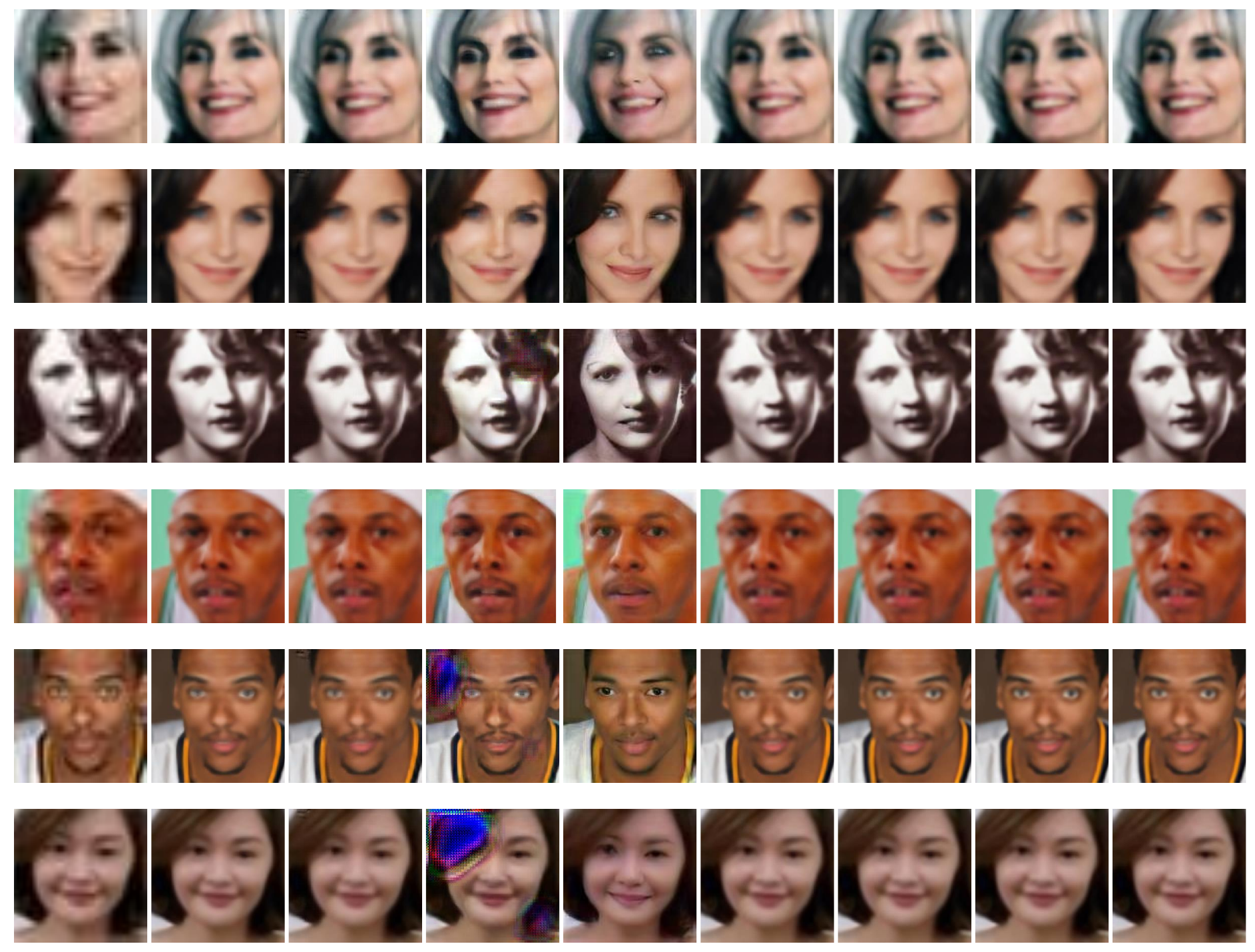}
  \vspace{-4mm}
  \caption{\textbf{Face hallucination results on the real-world images.}. From left to right: results of bicubic, DICNet, DICGAN, WaveletSR, HiFaceGAN, EDSR, RDN, RCAN and HAN.}
  \vspace{-4mm}
  \label{fig:real} 
\end{figure*}


\subsection{The Statistics of EDFace-Celeb-1M}

Lastly, we present the statistics of specific attributes of the developed EDFace-Celeb-1M dataset as follows.

\begin{itemize}
\item
First, we show some representative high-resolution face images from the EDFace-Celeb-1M dataset in Figure \ref{representative_image}. It is obvious that our dataset includes different races including White, Black, Asian and Latino, from different countries. In addition, these faces exhibit evident pose and appearance variations (e.g. glasses). Figure \ref{face_pose} presents the statistics of the face pose.
\item
Second, we demonstrate the ratios of races and gender. $64\%$ of the images of our dataset are from males and the rest $36\%$ images are from females. The ratios of White, Black, Asian and Latino are about $31.1\%$, $19.2\%$, $19.6\%$ and $18.3\%$, respectively, which are relatively balanced compared with existing datasets of face images.

\item 
Third, we also analyze the distribution of age of the celebrities we include in our dataset, and show the results in Figure \ref{age}. The age of celebrities is estimated by the model from Insightface. Roughly, the majority of age is between $25$ and $55$, which aligns well with the property of demography. Notably, our dataset includes also celebrities younger than $20$ and older than $60$. 

\item
Fourth, Figure \ref{resolution} presents the statistics of the resolution of face images before the resizing operation. In general, all the ground-truth high-resolution images are resized from images with resolution greater than $128 \times 128$.

\item
Fifth, the proposed dataset provides fixed training and testing subsets. Each subsets includes high-quality images and corresponding low-quality images. All images are labeled as HR and LR with factors( \textit{e.g.}, $2\times$, $4\times$, $4\times\_BD$, $4\times\_DN$ and $8\times$). Based on the labels and fixed subsets, the dataset can make the reproducibility and the fair comparison easier.

\end{itemize}

\section{Experiments}
\label{sec:experiment}

In this section, we introduce the evaluation protocols and benchmark the existing face hallucination methods and representative SISR methods on the proposed EDFace-Celeb-1M dataset.

\subsection{Evaluated Methods}

In this benchmark study, we investigate four state-of-the-art face hallucination methods (\textit{i.e.}, Deep Iterative Collaboration Network (DICNet) \cite{ma2020deep}, Deep Iterative Collaboration GAN (DICGAN) \cite{ma2020deep}, Wavelet-based Super-Resolution Network (WaveletSRNet) \cite{huang2017wavelet} and HiFaceGAN \cite{yang2020hifacegan}),  and four SISR methods (\textit{i.e.}, EDSR \cite{lim2017enhanced}, Holistic Attention Network (HAN) \cite{niu2020single}, Residual Dense Network (RDN) \cite{zhang2018residual}, and Residual Channel Attention Network (RCAN) \cite{zhang2018image}). All methods are based on deep learning. Specifically, DICGAN \cite{ma2020deep} is an iterative framework of recurrently estimating landmarks and recovering high-resolution images. Wavelet coefficients are introduced in WaveletSRNet \cite{huang2017wavelet} to deal with very-low-resolution facial images. HiFaceGAN \cite{yang2020hifacegan} is proposed to formulate the face restoration task as a generation problem guided by semantics, and this problem is addressed by a multi-stage framework containing several units of collaborative suppression and replenishment. For the SISR task, EDSR is an enhanced deep super-resolution network containing several Resblocks. HAN is proposed in \cite{niu2020single} to model the correlation among different convolution layers with a layer attention module and channel-spatial attention module. RDN \cite{zhang2018residual} is a residual dense network to exploit the hierarchical features from both the local and global perspectives. RCAN \cite{zhang2018image} is a deep residual channel attention network with both short and long skip connections. Channel attention is also adopted in this network for better performance. In summary, we select the above eight methods for three criteria. First, these methods achieve high values of the commonly used metrics like PSNR and SSIM. Second, the codes of these methods are publicly available. Third, these methods are proposed for face hallucination or image super-resolution.

\subsection{Implementation Details}

Our dataset has four different degradation settings. Each setting corresponds to pairs of low-resolution and high-resolution face images, which are used to train different models. We use the code released from the original publications. For fair comparisons, the learning rate and epoch number for all methods are set as $0.0001$ and $20$, respectively. All models are trained using V100 GPUs. We conduct the calculation of different metrics in the RGB space to access the results. During the training stage, all models are trained on the training subset and the testing subset is not used. After training, we evaluate the models on the testing subset.

\subsection{Face Super-Resolution}

To evaluate the four methods on face hallucination, we provide the quantitative results of PSNR and SSIM on the EDFace-Celeb-1M dataset in Table \ref{tab:psnr_ssim}. Based on the PSNR and SSIM values, DICNet achieves the best performance on $4\times$, $4\times\_{BD}$, $4\times\_{DN}$ and $8\times$ degradation settings. RCAN achieves the best performance on $2\times$. In terms of SSIM, the best performance values on $4\times$, $4\times\_{BD}$, $4\times\_{DN}$, and $8\times$ are also obtained by WaveletSR. The Table \ref{tab:psnr_ssim} does not provide the results of DICNet and DICGAN on $X2$ setting because the two methods do not work on this setting. Figures \ref{fig:results_8x}, \ref{fig:results_4x}, \ref{fig:results_bd} and \ref{fig:results_dn} show the visual comparison of different methods on the EDFace-Celeb-1M dataset.

\subsection{Face Alignment}

Face alignment \cite{albiero2021img2pose} is an important task in the field of computer vision. Apart from the PSNR and SSIM, we also evaluate the performance of face hallucination methods via face alignment. Specially, we first use a popularly used alignment model from Dlib to estimate the 68 landmarks of high-resolution face images, and the super-resolved face images generated via different methods. Then, we use a metric commonly employed in face alignment, Normalized Root Mean Squared Error (NRMSE), to evaluate the error between the landmarks from the HR (High-Resolution) and SR (Super-Resolution) face images. A smaller NRMSE value indicates better face alignment performance, which corresponds to a better face hallucination method. We can see from Table \ref{tab:alignment_parsing_id} that the best performance on $2\times$, $4\times$, $4\times\_{BD}$, $4\times\_{DN}$ and $8\times$ are obtained by HAN, HAN, RCAN, HAN and RDN, respectively. 

We have an interesting observation that image SR methods achieve better performance than face hallucination methods regarding the metric of face alignment. This is because image SR methods mainly consider to super-resolve the LR images and ignore the adjustment of high-level information (like landmarks). Therefore, these methods can maintain the original landmarks better. For face hallucination methods, they usually consider the high-level information to help update the super-resolving process, which may cause the change of landmark.

\begin{figure}
  \centering
\includegraphics[width=0.9\linewidth ]{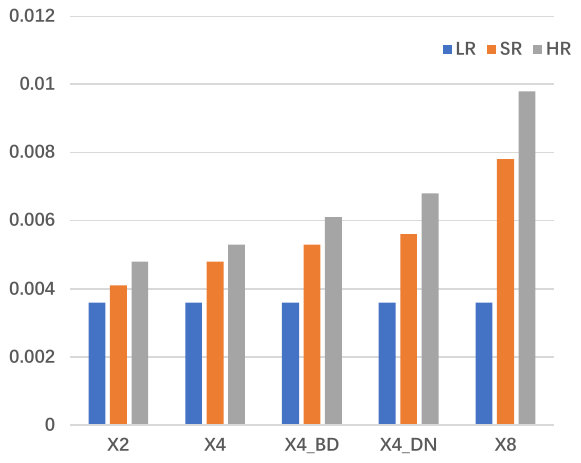}
\caption{{\bf An increase in similarity across different people after the hallucination.}}
\label{face_id_different}
\end{figure}

\subsection{Face Identity Information}

Facial identity information is important during face hallucination. An ideal face hallucination approach should ensure that the ID information of HR face images and SR face images are the same. However, in the real world, it is difficult for the face hallucination methods to generate SR results that are the same as HR images. In this section, we use a popular face ID information extractor \cite{deng2019arcface} to extract ID information from HR face images and SR face images generated by different face hallucination methods. We then calculate the cosine distance between them. The larger value indicates that the ID information loss is less, which corresponds to a better face hallucination method. Table \ref{tab:alignment_parsing_id} shows the results of different models. We can see that the best performance on $2\times$, $4\times$, $4\times\_{BD}$, $4\times\_{DN}$ and $8\times$ are obtained by HAN \& WaveletSR, RCAN, RCAN, RCAM and RCAN, respectively.

Based on results from Table \ref{tab:alignment_parsing_id}, we can find that face hallucination methods do not show better performance than image SR methods regarding the metric of face identification. We suspect that both of the two kinds methods do not consider the loss functions about identification information during the training stage. In the future, it may be a meaningful direction for researchers to study face identification information for the task of face hallucination. Finally, we extract similarity scores across different people's images before hallucination, and compare that with extracted similarity scores across different people's images after hallucination (by RCAN method) and their high-resolution versions. The Figure \ref{face_id_different} shows the similarity.

\subsection{Comparison on the Real LR Face Images}

In addition, we also show the performance of the evaluated methods in the case of real-world scenarios based on our EDFace-Celeb-1M dataset. Taking a real-world low-resolution face image from our proposed dataset, we process it by different methods to generate SR face images, and the results are shown in Figure \ref{fig:real}. We can find that most of the current face hallucination methods can improve the quality of low-resolution images.

\begin{figure}
  \centering
  \includegraphics[width=0.9 \linewidth ]{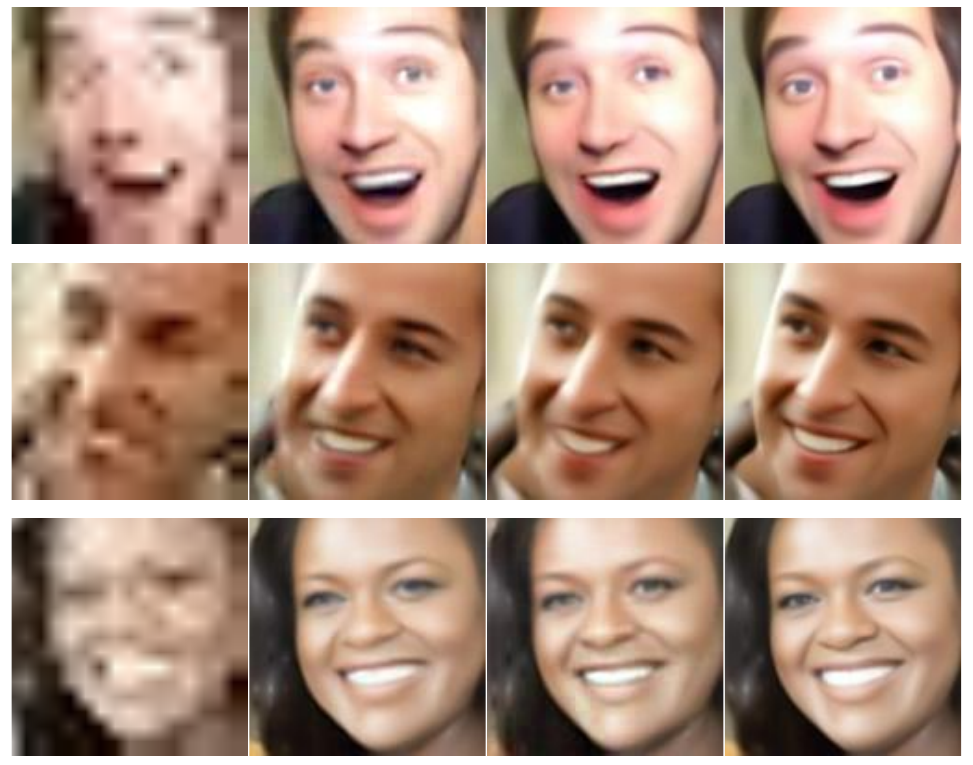}
  \caption{\textbf{Face hallucination results trained on different numbers of training samples.} From left to right, the columns show results of the DICNet trained using $0.3M$, $0.7M$ and $1.35M$ training samples from the EDFace-Celeb-1M dataset.}
  \label{fig:different_samples} 
\end{figure}

\begin{table}
  \centering 
    \caption{\textbf{Performance comparison of DICNet \cite{ma2020deep} trained using different numbers of training samples from the EDFace-Celeb-1M dataset}. Results are reported in terms of PSNR and SSIM. $0.3M$, $0.7M$ and $1.35M$ represent the size of training samples. The highest, second highest and lowest results are highlighted in bolded, blue and red, respectively.}
    \begin{tabular}{c | c | c c c c  }
    \hline
    Scale &  Metrics  &  $0.3M$ & $0.7M$ & $1.35M$  \\
    \hline
    \multirow{2}{*}{$4\times$} & PSNR & \textcolor{red}{28.42}  & \textcolor{blue}{28.78} & \textbf{29.06}  \\
                     & SSIM & \textcolor{red}{0.8265} & \textcolor{blue}{0.8362} & \textbf{0.8453}  \\
    \hline
    \multirow{2}{*}{$4\times\_BD$} & PSNR & \textcolor{red}{28.93}  & \textcolor{blue}{29.36} & \textbf{29.68}  \\
                     & SSIM & \textcolor{red}{0.8053} & \textcolor{blue}{0.8162} & \textbf{0.8213}  \\
    \hline
    \multirow{2}{*}{$4\times\_DN$} & PSNR & \textcolor{red}{27.23}  & \textcolor{blue}{27.68} & \textbf{27.96}  \\
                     & SSIM & \textcolor{red}{0.7986} & \textcolor{blue}{0.8073} & \textbf{0.8117}  \\
    \hline
    \multirow{2}{*}{$8\times$} & PSNR & \textcolor{red}{24.63} & \textcolor{blue}{24.95} & \textbf{25.29}  \\
                     & SSIM & \textcolor{red}{0.7216} & \textcolor{blue}{0.7376} & \textbf{0.7453} \\
    \bottomrule
    \end{tabular}
    \label{tab:different_samples}
\end{table}%

\subsection{Impacts of Training Samples}

Given that the proposed EDFace-Celeb-1M is a large-scale public face hallucination dataset, we conduct an experimental study to explore the effect of the size of training samples for face hallucination. Figure \ref{fig:different_samples} and Table \ref{tab:different_samples} show that models can achieve better performance with an increasing number of training samples.

\section{Discussions}

In this section, we discuss the quality of the dataset and the effects we have made to fairly evaluate the current methods.

\subsection{The Quality of the Dataset}

In summary, we provide an important dataset for the face hallucination community. The literature currently lacks a dataset specific to the face hallucination task and the proposed dataset makes the reproducibility and the fair comparison much easier for further research. To ensure the quality of this dataset, we make several efforts. First, to improve the quality of facial images in our dataset, we use a face detector to obtain facial images. In addition, we also manually remove some low-quality facial images. Second, to improve models' reproducibility on the dataset, we fixed the two training and testing subsets. As a comparison, previous methods randomly choose the two subsets. Third, different from previous methods which only provide one or two degraded settings, the provided dataset provides five degraded settings. In this way, the proposed dataset can better evaluate the performance of different face hallucination methods. Fourth, different from previous methods which ignore the problem of ethnicity, the proposed dataset provides a relatively balanced race composition. Fifth, different from some previous methods which evaluate their methods on relatively small-scale datasets, the provided dataset is the largest publicly available face hallucination dataset. In future, we will consider building more datasets to benefit the development of face hallucination, including using stronger face detectors to better detect faces and providing video sequences to help learn spatio-temporal information.

\subsection{Fairness}

In summary, to fairly evaluate the current methods, this paper makes some significant efforts. First, we build a large-scale publicly available dataset and fix the training and testing sets to conduct experiments, rather than randomly choose samples. Second, we use the popular metrics including PSNR and SSIM to compare different methods. Third, to further evaluate the current methods, we design two task-driven ablation studies including landmark detection and identify preservation.

\section{Conclusions}
\label{sec:conclusion}

In this paper, we first propose the largest publicly available face hallucination dataset with relatively balanced race composition. It contains $1.5$ million pairs of LR and HR face images for training and testing, and $140$K real-world tiny face images for quantitative comparisons. Thanks to the proposed EDFace-Celeb-1M dataset, the following face hallucination can evaluate their methods on a public and fixed division of training and testing samples, which significantly makes the comparison convenient and improves the reliability.

In addition, given that the current face hallucination methods are evaluated on privately synthesized datasets, we \textbf{benchmark} four public available face hallucination methods, and four SISR methods on the proposed EDFace-Celeb-1M dataset. The proposed dataset will be made publicly available to encourage the development of face hallucination algorithms.


\bibliographystyle{IEEEtran}
\bibliography{egbib_new}

\begin{IEEEbiography}[{\includegraphics[width=1in,height=1.25in,clip,keepaspectratio]{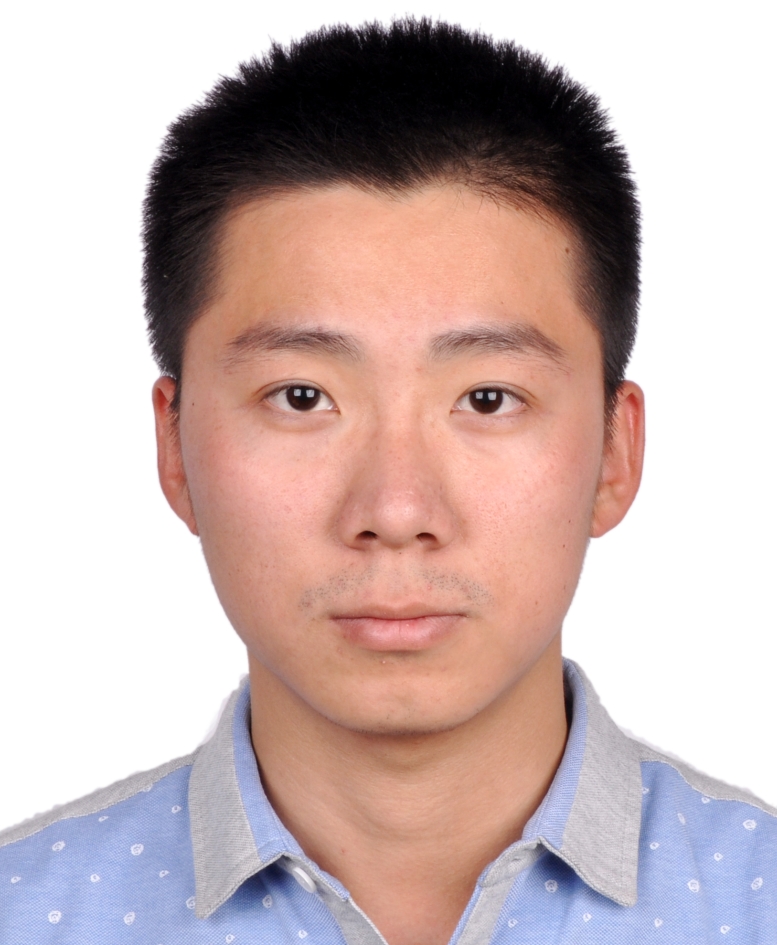}}]{Kaihao Zhang}
is currently pursuing the Ph.D. degree with the College of Engineering and Computer Science, The Australian National University, Canberra, ACT, Australia. His research interests focus on computer vision and deep learning. He has more than 30 referred publications in international conferences and journals, including CVPR, ICCV, ECCV, NeurIPS, AAAI, ACMMM, TPAMI, IJCV, TIP, TMM, etc. 
\end{IEEEbiography}

\begin{IEEEbiography}[{\includegraphics[width=1in,height=1.25in,clip,keepaspectratio]{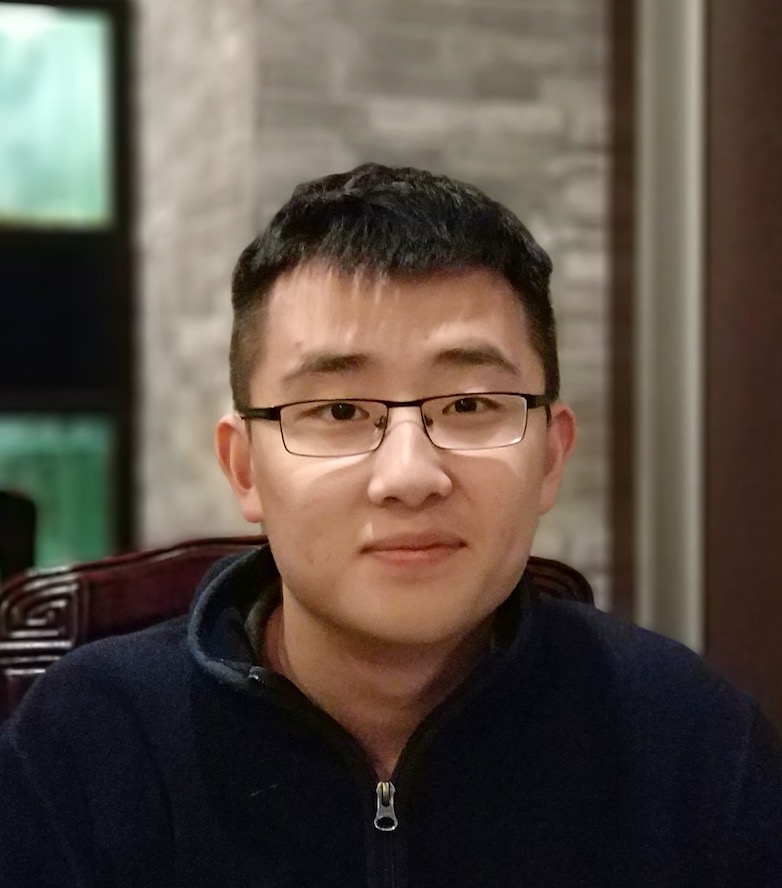}}]{Dongxu Li}
is a Ph.D candidate at The Australian National University. His research interests are mainly computer vision and deep learning, including visual sequence representation learning, vision-language learning and multi-modal learning. Before starting PhD, Dongxu obtained his Bachelor degree from The Australian National University with first-class honours in Computing.
\end{IEEEbiography}

\begin{IEEEbiography}[{\includegraphics[width=1in,height=1.25in,clip,keepaspectratio]{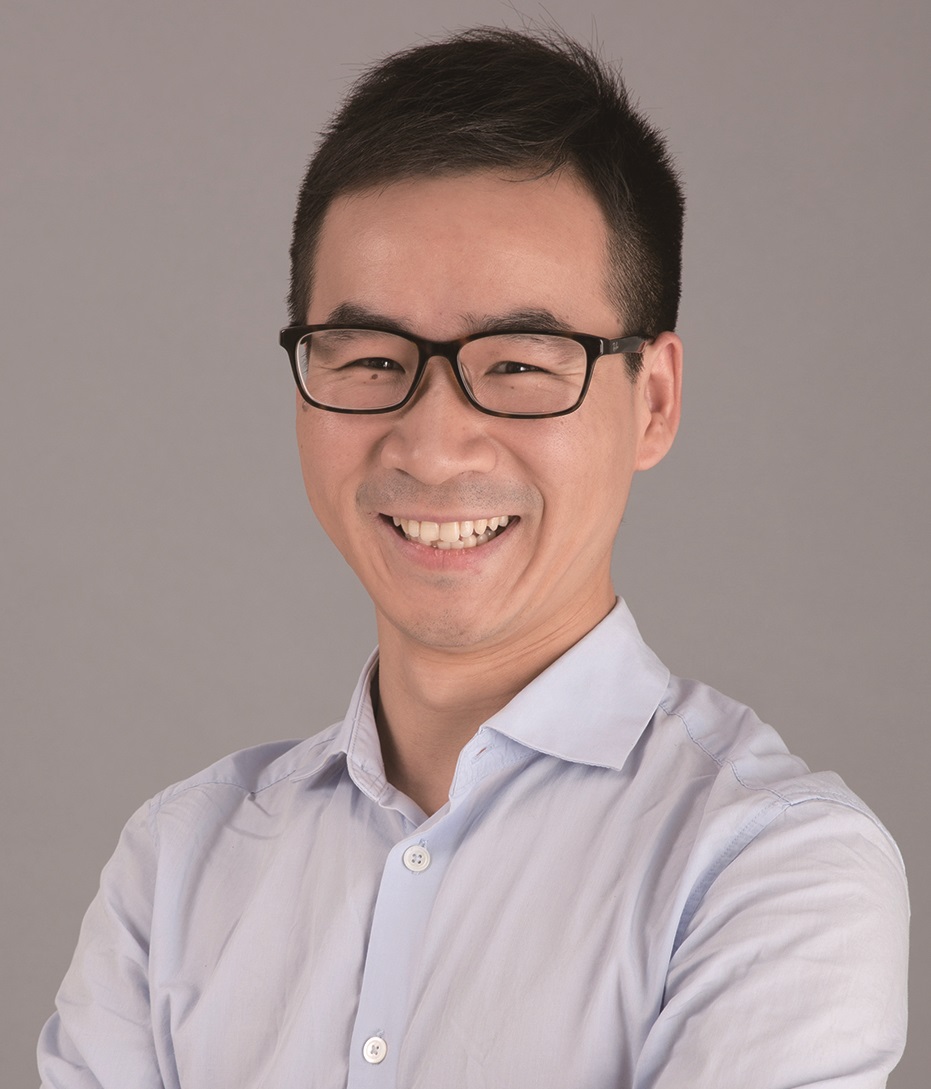}}]{Wenhan Luo}
is currently an Associate Professor in Sun Yat-sen University. Prior to that, he worked as a research scientist for Tencent and Amazon. He has published over 40 papers in top conferences and leading journals, including ICML, CVPR, ICCV, ECCV, ACL, AAAI, ICLR, TPAMI, IJCV, TIP, etc. He also has been reviewer, senior PC member and Guest Editor for several prestigious journals and conferences. His research interests include several topics in computer vision and machine learning, such as image/video synthesis, image/video quality restoration, reinforcement learning. He received the Ph.D. degree from Imperial College London, UK, 2016, M.E. degree from Institute of Automation, Chinese Academy of Sciences, China, 2012 and B.E. degree from Huazhong University of Science and Technology, China, 2009.
\end{IEEEbiography}

\begin{IEEEbiography}[{\includegraphics[width=1in,height=1.25in,clip,keepaspectratio]{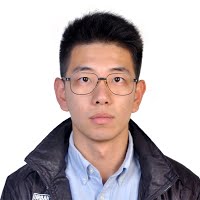}}]{Jingyu Liu}
is currently a research engineer in Tencent. His research interests include object recognition and multi-modal learning. He obtained his PhD degree from Institute of Automation, Chinese Academy of Science in 2018.
\end{IEEEbiography}

\begin{IEEEbiography}[{\includegraphics[width=1in,height=1.25in,clip,keepaspectratio]{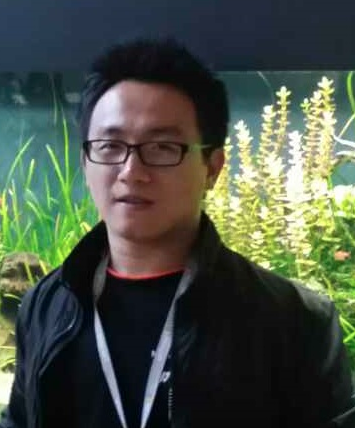}}]{Jiankang Deng} 
obtained his PhD degree from Imperial College London (ICL), supervised by Prof. Stefanos Zafeiriou and funded by the Imperial President's PhD Scholarships. His research topic is deep learning-based face analysis, including detection, alignment, reconstruction, recognition and generation etc. He is a reviewer in prestigious computer vision journals and conferences including T-PAMI, IJCV, CVPR, ICCV and ECCV. He is one of the main contributors to the widely used open-source platform Insightface. He is a student member of the IEEE.
\end{IEEEbiography}

\begin{IEEEbiography}[{\includegraphics[width=1in,height=1.25in,clip,keepaspectratio]{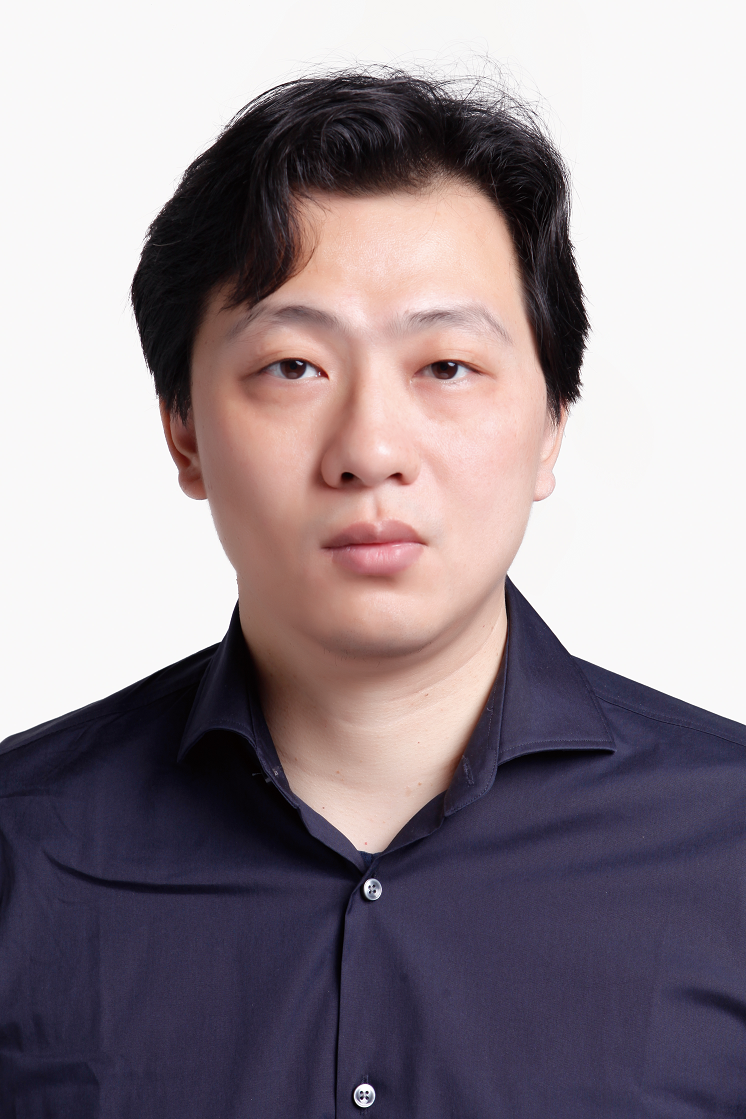}}]{Wei Liu}
is currently a Distinguished Scientist of Tencent AI Lab and a director of Computer Vision Center. Prior to that, he received the Ph.D. degree in EECS from Columbia University, New York, NY, USA, and was a research scientist of IBM T. J. Watson Research Center, Yorktown Heights, NY, USA. Dr. Liu has long been devoted to research and development in the fields of machine learning, computer vision, information retrieval, big data, etc. Till now, he has published more than 150 peer-reviewed journal and conference papers, including Proceedings of the IEEE, TPAMI, TKDE, IJCV, NIPS, ICML, CVPR, ICCV, ECCV, KDD, SIGIR, SIGCHI, WWW, IJCAI, AAAI, etc. His research works win a number of awards and honors, such as the 2011 Facebook Fellowship, the 2013 Jury Award for best thesis of Columbia University, the 2016 and 2017 SIGIR Best Paper Award Honorable Mentions, and the 2018 "AI's 10 To Watch" honor. Dr. Liu currently serves as an Associate Editor to several international leading AI journals and an Area Chair to several international top-tier AI conferences, respectively.

\end{IEEEbiography}

\begin{IEEEbiography}[{\includegraphics[width=1in,height=1.25in,clip,keepaspectratio]{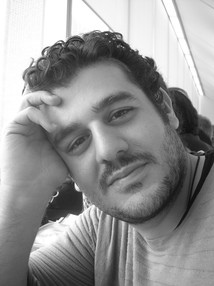}}]{Stefanos Zafeiriou} is currently a Professor in Machine Learning and Computer Vision with the Department of Computing, Imperial College London, London, U.K, and an EPSRC Early Career Research Fellow. He served Associate Editor and Guest Editor in various journals including TPAMI, IJCV, TAC, CVIU,
and IVC. He has been a Guest Editor of 8+ journal special issues and co-organised over 16 workshops/special sessions on specialised computer vision topics in top venues. He has co-authored 70+ journal papers mainly on novel statistical machine learning methodologies applied to computer vision problems, such as 2-D/3-D face analysis, deformable object fitting and tracking, shape from shading, and human behaviour analysis, published in the most prestigious journals in his field of research, such as TPAMI, IJCV, and many papers in top conferences, such as CVPR, ICCV, ECCV, ICML. His students are frequent recipients of very prestigious and highly competitive fellowships, such as the Google Fellowship x2, the Intel Fellowship, and the Qualcomm Fellowship x4. He has more than 20K+ citations to his work, h-index 64. He was the General Chair of BMVC 2017. He is a member of the IEEE.
\end{IEEEbiography}

\end{document}